\documentclass[runningheads]{llncs}
\usepackage[T1]{fontenc}
\usepackage{algorithm}
\usepackage{algpseudocode}
\usepackage{amsmath}
\usepackage{amssymb}
\usepackage{bm}
\usepackage{booktabs}
\usepackage{color}
\usepackage{comment}
\usepackage{epsfig}
\usepackage{graphicx}
\usepackage[hidelinks]{hyperref}
\usepackage{multirow}
\usepackage{orcidlink}
\usepackage{subfig}
\usepackage{times}

\graphicspath{{images/}}

\newcommand{\eg}{e.g.}
\newcommand{\etal}{et al. }

\begin{document}

\title{Energy-Based Open-Set Active\\ Learning for Object Classification}

\author{Zongyao Lyu\orcidlink{0000-0002-7542-5818} \and
William J. Beksi\orcidlink{0000-0001-5377-2627}}

\institute{The University of Texas at Arlington, Arlington TX 76019, USA}

\maketitle              

\begin{abstract}
Active learning (AL) has emerged as a crucial methodology for minimizing
labeling costs in deep learning by selecting the most valuable samples from a
pool of unlabeled data for annotation. Traditional AL operates under a
closed-set assumption, where all classes in the dataset are known and
consistent. However, real-world scenarios often present open-set conditions in
which unlabeled data contains both known and unknown classes.  In such
environments, standard AL techniques struggle. They can mistakenly query
samples from unknown categories, leading to inefficient use of annotation
budgets. In this paper, we propose a novel dual-stage energy-based framework
for open-set AL. Our method employs two specialized energy-based models (EBMs).
The first, an energy-based known/unknown separator, filters out samples likely
to belong to unknown classes. The second, an energy-based sample scorer,
assesses the informativeness of the filtered known samples. Using the energy
landscape, our models distinguish between data points from known and unknown
classes in the unlabeled pool by assigning lower energy to known samples and
higher energy to unknown samples, ensuring that only samples from classes of
interest are selected for labeling. By integrating these components, our
approach ensures efficient and targeted sample selection, maximizing learning
impact in each iteration. Experiments on 2D (CIFAR-10, CIFAR-100, TinyImageNet)
and 3D (ModelNet40) object classification benchmarks demonstrates that our
framework outperforms existing approaches, achieving superior annotation
efficiency and classification performance in open-set environments.


\end{abstract}
\vspace{-4mm}

\section{Introduction}
\label{sec:introduction}
In recent years, deep learning has driven unprecedented advancements in
computer vision tasks including semantic segmentation, object detection and
classification, and much more. These achievements have been fueled by the
availability of labeled large-scale datasets. Yet, the annotation process
requires substantial time, effort, and expertise, especially for tasks
involving 3D data due to the complexity of geometric representations.  To
mitigate this burden, the field has increasingly turned to AL, a strategy that
aims to reduce labeling costs by selectively querying the most valuable samples
from an unlabeled pool \cite{cohn1996active,settles2009active,top2011active}.

\begin{figure}[t]
\centering
\includegraphics[width=0.8\textwidth]{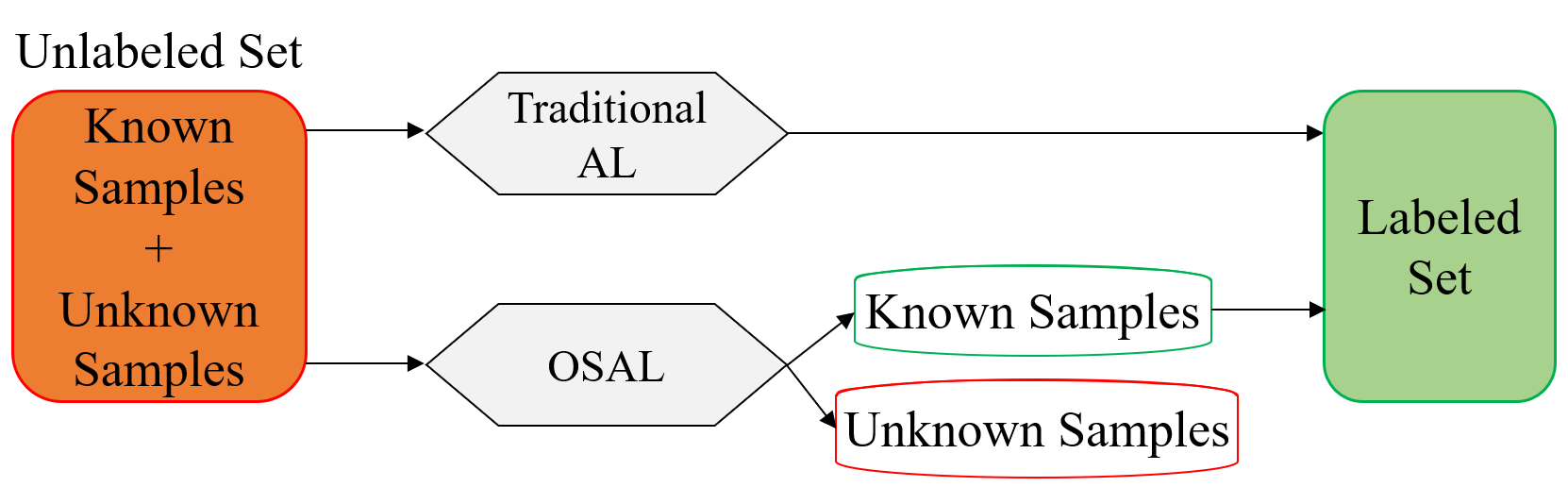}
\caption{An illustration of the OSAL problem.  The unlabeled set contains
samples from both known and unknown classes.  Traditional AL applies selection
strategies indiscriminately, often allocating a labeling budget to unknown
samples. In contrast, OSAL first filters out unknown samples and retains only
likely known samples. A standard AL strategy is then applied to the filtered
set to select informative samples for annotation, improving labeling efficiency
under open-set conditions.}
\vspace{-3mm}
\label{fig:osal_overview}
\end{figure}

The vast majority of AL methods have been developed under a closed-set
assumption, where both the labeled and unlabeled datasets consist exclusively
of a predefined set of known classes. These methods typically rely on metrics
such as uncertainty and diversity to identify samples that lie near decision
boundaries, where additional labels are most likely to improve model accuracy.
Nonetheless, in real-world scenarios the assumption of a closed-set environment
often fails. Unlabeled datasets may contain classes not in the labeled training
set, creating an open-set problem.

Open-set conditions pose significant challenges to traditional AL approaches.
For example, the model may encounter samples from unknown categories, resulting
in ambiguous predictions with high uncertainty. In real-world applications,
unlabeled data often includes such unknown classes not represented in the
labeled dataset. Without mechanisms to distinguish between known and unknown
categories, traditional AL strategies risk querying samples from these unknown
classes, leading to inefficient use of annotation resources and potentially
degrading model performance. This scenario, known as open-set AL (OSAL), demands
new strategies that can accurately differentiate between known and unknown
categories, ensuring only relevant and useful samples are selected for annotation
(Fig.~\ref{fig:osal_overview}).

To address the challenges of OSAL, we propose an energy-based OSAL (EB-OSAL)
framework built on two specialized EBMs in a dual-stage design. The system
consists of two key components: the energy-based known/unknown separator (EKUS)
and the energy-based sample scorer (ESS). EKUS filters out samples that likely
belong to unknown classes, allowing the model to concentrate on data relevant to
the target task. Then, ESS ranks the informativeness of the remaining ``likely
known'' samples, selecting those that best refine the decision boundaries between
known categories. By utilizing the energy landscape of EBMs to model the
likelihood of samples belonging to known or unknown classes, our approach
effectively handles the separation and selection of informative samples.

\textit{We validate EB-OSAL on 2D object classification benchmarks and extend
our framework to the task of 3D object classification, an underexplored yet
important domain where annotation cost is even higher due to the complexity of
the data, and demonstrate its versatility across modalities.} In summary, our
contributions are the following.
\begin{enumerate}
  \item A novel dual-stage energy-based framework for OSAL that efficiently
  handles both known and unknown samples in unlabeled data while accurately
  selecting the most informative known samples for annotation.
  \item An EKUS component that effectively filters out samples likely belonging
  to unknown classes, allowing the model to focus on relevant data for the
  target task.
  \item An ESS component that accurately ranks the informativeness of known
  samples, ensuring that the most valuable data points are selected for
  annotation.
  \item New benchmark results for OSAL on the CIFAR-10, CIFAR-100, TinyImageNet,
  and ModelNet40 datasets, demonstrating the superiority of our approach for
  both 2D and 3D classification tasks in open-set environments.
\end{enumerate}
Our source code is publicly available at \cite{eb-osal}.

\section{Related Work}
\label{sec:related_work}
\subsection{Open-Set Recognition}
Open-set recognition (OSR) aims to identify whether a sample belongs to a known
class or an unseen category. Early work used statistical methods, such as extreme
value theory, to model class-distribution tails and reject unknowns (\eg,
\cite{scheirer2012toward,scheirer2014probability,scheirer2017extreme}). Later,
neural network approaches (\eg, OpenMax \cite{bendale2016towards}, MetaMax
\cite{lyu2023metamax}) modified classifier outputs to handle unknowns, while
reconstruction-based models detect anomalies via reconstruction errors
\cite{yoshihashi2019classification}. Other works use generative models to
produce synthetic samples, improving discrimination between known and unknown
classes \cite{ge2017generative,kong2021opengan}.

In the 3D domain, Cen \etal \cite{cen2021open} propose a LiDAR-based open-set
object detection formulation, Alliegro \etal \cite{alliegro20223dos} introduce
the 3DOS benchmark for open-set point cloud classification, and Li and Dong
\cite{li2023open} develop an adversarial prototype method for open-set 3D
semantic segmentation. These studies highlight that open-set challenges also
arise in 3D domains and require tailored solutions.

\subsection{Active Learning}
AL reduces labeling costs by selecting the most informative samples for annotation,
maximizing performance under limited budgets. AL proceeds iteratively, training
on the labeled set and querying an oracle (e.g., a human annotator) to label new
samples from an unlabeled pool. In pool-based AL, strategies are commonly
categorized as uncertainty-based, diversity-based, or hybrid.

Uncertainty-based methods query samples the model is least confident about
\cite{joshi2009multi,gal2017deep,beluch2018power}, while diversity-based
methods, such as core-set, select samples covering a broad data range
\cite{nguyen2004active,sener2018active}. Hybrid methods combine both criteria
to overcome the weaknesses of single strategies
\cite{huang2014active,yang2015multi,ash2020deep,yan2022clustering,lyu2024semi}.

Recent work has extended AL to 3D point clouds. For instance, Wu \etal
\cite{wu2021redal} develop ReDAL for efficient segmentation. SSDR-AL by Shao
\etal \cite{shao2022active} selects informative regions via spatial-structural
diversity. Xu \etal \cite{xu2023hierarchical} introduce HPAL with hierarchical
uncertainty and redundancy-aware sampling. Using image priors, Samet \etal
\cite{samet2023you} propose SeedAL. Annotator by Xie \etal
\cite{xie2023annotator} is a general 3D LiDAR segmentation baseline.

Despite this progress, these methods assume a closed-set scenario, where all
unlabeled samples belong to known classes. In open-set settings, encountering
unknown classes makes them less effective: uncertain predictions from unknowns
can be misinterpreted, thus wasting the annotation budget on samples irrelevant
to the target task.

\subsection{Open-Set Active Learning}
OSAL extends AL to settings where the goal is not only to query informative
samples, but also to filter out unknowns. Early approaches combined OSR
techniques with traditional AL strategies such as LfOSA \cite{ning2022active},
which models max activations to separate known and unknowns before querying,
and MQ-Net \cite{park2022meta}, which uses meta-learning to balance
informativeness and sample purity.

Recent methods include EOAL \cite{safaei2024entropic}, which employs two entropy
scores to separate known and unknowns, and 
BUAL \cite{zong2024bidirectional}, which introduces a bidirectional uncertainty
framework with random-label negative learning to suppress unknown samples while
querying informative known ones. In contrast, we propose a dual-stage EBM
framework that enhances informativeness and operates in both 2D and 3D domains.

Existing OSAL methods remain limited to 2D tasks. To our knowledge, no prior
work addresses OSAL for 3D object classification. \textit{Our work fills this
gap by extending energy-based OSAL to 3D, demonstrating its effectiveness in
high-dimensional, spatially structured tasks.}

\subsection{Energy-Based Models}
EBMs offer a flexible framework for modeling complex, high-dimensional data by
associating an energy value with each sample. Unlike traditional probabilistic
models that assign explicit probabilities, EBMs define an energy function
mapping input features to scalar values, with lower energy indicating a higher
likelihood of belonging to a specific distribution
\cite{lecun2006tutorial,grathwohl2019your}. The core idea is to shape the
energy landscape such that data similar to the training samples receives lower
energy, while outliers and unknowns are assigned higher energy
\cite{chen2023secure}. This makes EBMs well suited for OSR, outlier detection,
and uncertainty estimation.

In OSR, EBMs have proven effective in tasks such as out-of-distribution (OOD)
detection, anomaly detection, and domain adaptation \cite{xie2022active} by
separating known from unknown classes through energy values that reflect sample
confidence. For example, Zhai \etal \cite{zhai2016deep} show how energy can
serve as a criterion for anomaly detection by linking the energy landscape to
anomaly identification. Liu \etal \cite{liu2020energy} use energy scores for
OOD detection, outperforming softmax confidence by providing a more reliable
measure of input compatibility. Nevertheless, the potential of EBMs in AL under
open-set conditions where the unlabeled pool contains unseen categories remains
underexplored. We utilize EBMs to separate unknowns from knowns and to score
informative samples among the latter, thereby maximizing the impact of queried
data and enhancing robustness in open-set learning.

Beyond 2D images, EBMs have also been explored for 3D point clouds. Examples
include generative PointNet \cite{xie2021generative}, energy-based 3D object
detection \cite{gustafsson2021accurate}, and point cloud completion and
implicit representations \cite{cui2022energy,yamauchi2023optimizing}.
Differently, our framework is the first to apply EBMs to OSAL for 3D object
classification tasks.

\section{Preliminaries}
\label{sec:preliminaries}
EBMs are a class of generative models that represent probability distributions
using a scalar energy function $E_\theta(\mathbf{x})$. The function maps each
point $\mathbf{x}$ in the input space to an energy value, where lower energy
corresponds to a higher probability density. The probability distribution
$p_\theta(\mathbf{x})$ defined by an EBM takes the form
\begin{equation}
  p_\theta(\mathbf{x}) = \frac{\exp(-E_\theta(\mathbf{x}))}{Z(\theta)},
\end{equation}
where $Z(\theta) = \int_{\mathbf{x}} \exp(-E_\theta(\mathbf{x})) d\mathbf{x}$ is
the partition function, ensuring that $p_\theta(\mathbf{x})$ normalizes over the
entire input space. The computation of $Z(\theta)$ is often intractable in
high-dimensional spaces, making direct probability calculations challenging. As
a result, training methods focus on optimizing the energy function using
alternative approaches.

In practice, EBMs are trained by maximizing the likelihood of the observed data
distribution $p_{\text{data}}(\mathbf{x})$ \cite{du2019implicit}. This involves
updating the model parameters $\theta$ to minimize the Kullback-Leibler
divergence between $p_{\text{data}}(\mathbf{x})$ and $p_\theta(\mathbf{x})$.
The gradient of the log-likelihood $\mathcal{L}(\theta)$ with respect to
$\theta$ is
\begin{equation}
  \nabla_\theta \mathcal{L}(\theta) = \mathbb{E}_{\mathbf{x} \sim p_{\text{data}}} \left[ \nabla_\theta E_\theta(\mathbf{x}) \right] - \mathbb{E}_{\mathbf{x}^- \sim p_\theta} \left[ \nabla_\theta E_\theta(\mathbf{x}^-) \right],
\end{equation}
where $\mathbf{x}^-$ denotes negative samples drawn from the model distribution
$p_\theta(\mathbf{x})$.

Sampling from $p_\theta(\mathbf{x})$ to compute these expectations is often
achieved using Monte Carlo methods such as Langevin Monte Carlo or stochastic
gradient Langevin dynamics \cite{welling2011bayesian},
\begin{equation}
  \mathbf{x}^{(t+1)} = \mathbf{x}^{(t)} - \frac{\eta^2}{2} \nabla_{\mathbf{x}} E_\theta(\mathbf{x}^{(t)}) + \eta \epsilon^{(t)},
\end{equation}
where $\eta$ is the step size, and $\epsilon^{(t)} \sim \mathcal{N}(0, I)$
represents Gaussian noise. This stochastic process enables the model to explore
the energy landscape effectively and draw samples that approximate
$p_\theta(\mathbf{x})$.

The connection between EBMs and discriminative models lies in how energy
functions relate to logit outputs in classification. For a neural network
$f(\mathbf{x})$ mapping input $\mathbf{x}$ to logits $\mathbf{f}_i(\mathbf{x})$,
the energy function can be expressed as $E(\mathbf{x}, y) = -f_y(\mathbf{x})$
for a given class label $y$. The free energy $E(\mathbf{x})$ over all possible
labels $y$ is defined as
\begin{equation}
  E(\mathbf{x}) = -\log \sum_{y} \exp(f_y(\mathbf{x})).
\label{eq:free_energy}
\end{equation}
This formulation allows EBMs to be used in tasks such as OOD detection, where
samples with higher energy are identified as less likely to belong to the known
distribution.

\section{Method}
\label{sec:method}
\subsection{Problem Statement}
The goal of OSAL is to optimize the labeling process in the presence of unknown
classes. We are given a labeled dataset $D_L$, consisting of samples from known
classes $C_K$, and an unlabeled dataset $D_{UL}$ that contains both known-class
samples from $C_K$ and unknown-class samples from an unknown set $C_U$. The
primary challenge lies in separating relevant, informative samples from
irrelevant unknowns and querying the most valuable samples to improve the
model's performance on the known task.
To address this issue, we develop a dual-stage EBM framework. The first stage
filters out samples likely to belong to unknown classes, while the second stage
scores the remaining ``likely known'' samples based on informativeness, guiding
the sample-selection process for labeling. This formulation also applies to 3D
object classification where inputs are point clouds. In our 3D experiments, the
same OSAL protocol is used with point cloud inputs and 3D backbones as specified
below.

\begin{figure}
\centering
\includegraphics[width=0.9\textwidth]{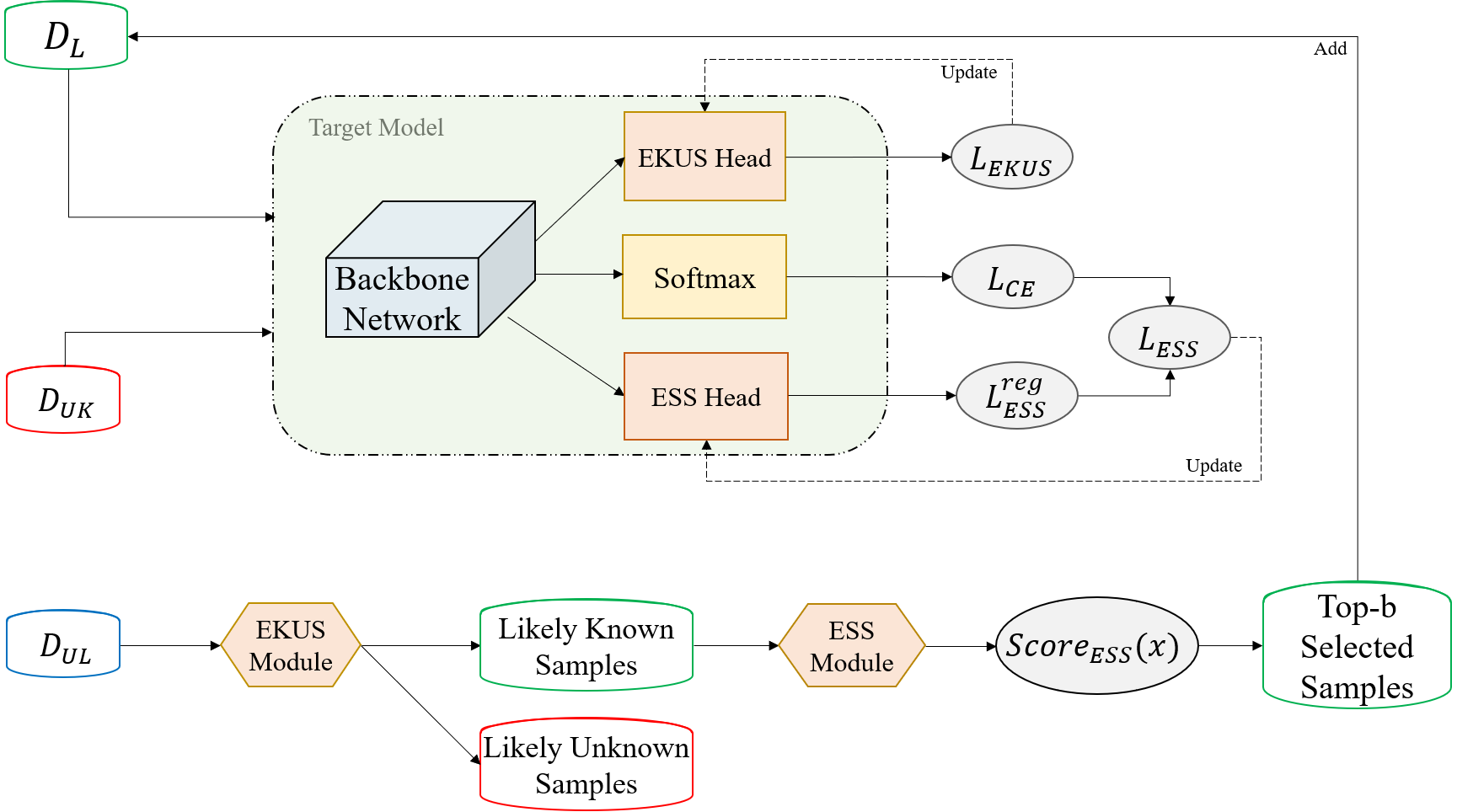}
\caption{ An overview of the EB-OSAL framework. The proposed approach follows a
dual-stage workflow that integrates EKUS and ESS for OSAL. The process operates
on labeled data $D_L$, containing known classes, and unlabeled data $D_{UL}$,
which includes both known and unknown samples. EKUS is trained to learn an
energy-based separation by anchoring known samples to low-energy regions and
identifying likely unknown samples within $D_{UL}$. At each AL cycle, EKUS
assigns energy scores to unlabeled samples by filtering out high-energy samples
as likely unknowns and retaining low-energy samples as likely knowns. The
retained samples are passed to ESS, which computes an informativeness score
based on uncertainty and energy. The top-ranked samples are selected for
annotation and added to $D_L$. EKUS and ESS are then retrained on the updated
labeled set, and the process iterates over multiple AL cycles.}
\label{fig:architecture_overview}
\vspace{-8mm}
\end{figure}

\subsection{Dual-Stage EBM Architecture Design}
The proposed framework utilizes two separate EBMs, EKUS and ESS, each designed
for a specific purpose. The training process for the dual-stage EBM involves
carefully crafted objectives to ensure optimal performance in their respective
roles.
Fig.~\ref{fig:architecture_overview} shows an overview of the architecture. For
2D object classification, we use a standard convolutional neural network
backbone (ResNet-18 \cite{he2016deep}). For 3D object classification, we use
PointNet \cite{qi2017pointnet} to handle unordered point sets, leveraging
shared multilayer perceptron (MLP) layers and symmetric pooling for permutation
invariance. 

\textbf{Energy-based known/unknown separator.} The EKUS module is designed to
filter out unknown samples from the unlabeled pool by learning an energy-based
separation between known and unknown data. Its architecture consists of a neural
network backbone that extracts feature representations, followed by an energy
head that maps each input to a scalar energy value. EKUS is trained such that
samples from known classes occupy low-energy regions, while unknown samples are
assigned higher energies. When EKUS is applied to score the unlabeled pool at
each AL cycle, samples with energy values exceeding a predefined threshold are
treated as unknown and excluded from the subsequent AL stage.

\textbf{Training EKUS.} EKUS is trained within the AL loop using labeled and
unlabeled data through a combination of energy anchoring, negative learning,
and pseudo-unknown margin enforcement. The objective is to learn a conservative
yet effective separator that progressively improves as more labeled data
becomes available.

At the beginning of the first AL cycle, EKUS is initialized using labeled and
unlabeled data without explicit unknown supervision. Labeled samples are used
to anchor the known-class energy manifold, while unlabeled samples are
regularized via negative learning \cite{kim2019nlnl}. Concretely, for an
unlabeled sample $x \in D_{UL}$, a complementary label $\bar{y}$ is randomly
sampled from the known class set $\mathcal{C}_K$, and the model is trained to
discourage confident assignment of $x$ to $\bar{y}$. This complementary
supervision prevents unlabeled samples from being confidently absorbed into
known classes and encourages higher energy values for samples that do not align
with the known-class structure.

As training proceeds, EKUS is used to identify a subset of unlabeled samples
that are likely to be unknown. Specifically, at each AL cycle we define an
unknown candidate set $D_{UK} \subset D_{UL}$ as the top-$\rho\%$ unlabeled
samples with the highest energy values under the current EKUS model. This set
is treated as a high-precision pseudo-unknown subset and is used to explicitly
reinforce separation through a margin-based loss. Formally, EKUS is optimized
using the objective
\begin{equation}
L_{\text{EKUS}} = L_{\text{hinge}} + \lambda L_{\text{contrastive}} + \gamma L_{\text{NL}},
\label{eq:ekus}
\end{equation}
where $\lambda$ and $\gamma$ are weighting hyperparameters controlling the
contributions of the contrastive and negative learning terms, respectively.

The hinge loss enforces explicit energy margins between known samples and
pseudo-unknown candidates,
\begin{equation}
    L_{\text{hinge}} =
    \sum_{x \in D_L} \bigl(\max(0, E_{\text{EKUS}}(x) - \delta_k)\bigr)^2 +
    \sum_{x' \in D_{UK}} \bigl(\max(0, \delta_u - E_{\text{EKUS}}(x'))\bigr)^2,
\label{eq:hinge}
\end{equation}
where $E_{\text{EKUS}}(x)$ denotes the energy computed by EKUS
using~\eqref{eq:free_energy}, and $\delta_k$ and $\delta_u$ are margin
thresholds for known samples and pseudo-unknown samples, respectively.  To
further encourage separation, a contrastive loss is applied:
\begin{equation}
    L_{\text{contrastive}}
    = \mathbb{E}_{x \sim \mathcal{D}_L,\; x' \sim \mathcal{D}_{UK}}
    \left[ E_{\text{EKUS}}(x) - E_{\text{EKUS}}(x') \right].
\label{eq:contrastive}
\end{equation}

Negative learning is incorporated as a regularization term over the unlabeled
pool,
\begin{equation}
    L_{\text{NL}}
    = \mathbb{E}_{x \sim \mathcal{D}_{UL},\; \bar{y} \sim \mathcal{U}(\mathcal{C}_K)}
    \left[ -\log \left( 1 - p_{\bar{y}}(x) \right) \right],
\label{eq:negative_learning}
\end{equation}
where $p_{\bar{y}}(x)$ denotes the predicted probability of class $\bar{y}$.
This term discourages confident assignment of unlabeled samples to known classes
and stabilizes the formation of the pseudo-unknown set $D_{UK}$ by reducing the
likelihood that known-but-unlabeled samples are incorrectly pushed into
high-energy regions.  Within the 3D backbone, the unordered point clouds are
processed via shared MLP layers to produce global features, which are then
passed to the energy head. The loss formulation is unchanged and operates
directly on the resulting 3D feature embeddings.

\textbf{Energy-based sample scorer.} Once EKUS has filtered out likely unknown
samples, ESS focuses on evaluating the informativeness of the remaining ``likely
known'' samples. ESS uses the same backbone as EKUS
for feature extraction to maintain consistency in feature representation. It
employs a separate linear layer to compute the energy score for each sample,
which now focuses on distinguishing samples near the decision boundaries of
known classes. Predictive uncertainty is calculated using a softmax function
applied to the model's outputs, with entropy serving as the measure of
uncertainty. ESS then combines the uncertainty and energy scores to determine
the overall informativeness of each sample, guiding the final selection for
annotation.

\textbf{Training ESS.} ESS focuses exclusively on samples filtered by EKUS and
trains to refine the decision boundaries between known classes. The overall
objective function for training ESS is given by
\begin{equation}
  L_{\text{ESS}} = L_{\text{CE}} + \alpha L_{\text{ESS}}^{\text{reg}},
\label{eq:ess}
\end{equation}
where the parameter $\alpha$ is a regularization coefficient that controls the
importance of energy regularization. The training process for ESS is centered on
a cross-entropy loss $L_{\text{CE}}$ for classification within the known classes,
combined with a regularization term to ensure that the energy for confidently
classified samples remains low. Additionally,
\begin{equation}
  L_{\text{ESS}}^{\text{reg}} = \sum_{x \in D_L} [\max(0, E_{\text{ESS}}(x) - \delta_s)]^2,
\end{equation}
where $E_{\text{ESS}}(x)$ represents the energy score computed by ESS for
sample $x$, and $\delta_s$ is a threshold that enforces a margin for the energy
values of known samples, ensuring a structured separation in the energy space.
This regularization term prevents high-energy values for well-understood samples,
focusing ESS on those near the decision boundaries.
In 3D, the entropy is computed over logits from the backbone-derived features.
The energy computation is identical to the 2D case, and the framework naturally
adapts to a 3D feature space.

\subsection{Sample Selection Strategy}
ESS plays a pivotal role in determining which ``likely known'' samples will
provide the most significant impact if labeled. It combines the energy score and
predictive uncertainty to assess each sample's informativeness. The scoring
function used is
\begin{equation}
  S_{\text{ESS}}(x) = U(x) + \beta \cdot E_{\text{ESS}}(x),
\label{eq:score}
\end{equation}
where $U(x) = - \sum_{i} P(y_i\,|\,x) \log P(y_i\,|\,x)$ represents the
uncertainty score derived from the entropy of the prediction of sample $x$ with
respect to label $y_i$, while $E_{\text{ESS}}(x)$ indicates the energy score
relative to known class boundaries. The weight $\beta$ is a hyperparameter that
balances the influence of uncertainty and proximity to decision boundaries.
Samples with the highest scores are prioritized, ensuring that the queried data
points will most effectively enhance the model's learning.

After selecting the top $b$ samples based on their scores, ESS updates the
labeled dataset and the training cycle is repeated. This iterative refinement
drives the model's performance, improving its capacity to handle known classes
accurately while maintaining robustness to unknown data. The same scoring
function is used for both 2D and 3D cases. The whole process is summarized in
Alg.~\ref{alg:method}.
\vspace{-2mm}

\begin{algorithm}
\caption{Energy-Based Open-Set Active Learning}
\begin{algorithmic}[1]
\Require
    Labeled data $D_L$, unlabeled data $D_{UL}$;
    Number of AL cycles $C$, annotation budget $b$;
    margins $\delta_k, \delta_u, \delta_s$;
    regularization weights $\lambda, \gamma, \alpha, \beta$;
    top-ratio $\rho$
\Ensure
    Updated labeled dataset $D_L$

\State Initialize EKUS and ESS modules

\For{$c = 1, \dots, C$}

    \State \textbf{Train EKUS} on $D_L \cup D_{UL}$:
    \State Compute energies $E_{\text{EKUS}}(x)$ for $x \in D_L \cup D_{UL}$
    \State Define $D_{UK} \subset D_{UL}$ as top-$\rho\%$ highest-energy samples
    \State Minimize $L_{\text{EKUS}}$ using Eq.~\eqref{eq:ekus}

    \State \textbf{Train ESS} on $D_L$ by minimizing $L_{\text{ESS}}$ using~\eqref{eq:ess}

    \State $\text{likely\_known\_set} \leftarrow \{x \in D_{UL} \mid E_{\text{EKUS}}(x) < \delta_k\}$ \Comment{Filter likely known samples}

    \State \textbf{Score likely known samples}:
    \For{$x \in \text{likely\_known\_set}$}
        \State Compute $U(x)$, $E_{\text{ESS}}(x)$, and $S_{\text{ESS}}(x)$ using~\eqref{eq:score}
    \EndFor

    \State Select top-$b$ samples with highest scores from $\text{likely\_known\_set}$ and add to $D_L$
    \State Remove selected samples from $D_{UL}$

\EndFor
\end{algorithmic}
\label{alg:method}
\end{algorithm}
\vspace{-7mm}

\section{Experiments}
\label{sec:experiments}
\subsection{Experimental Setup}
\textbf{Datasets.}
To evaluate the effectiveness of the EB-OSAL framework on 2D object
classification, we conducted experiments on three widely used benchmarks:
CIFAR-10, CIFAR-100 \cite{krizhevsky2009learning}, and TinyImageNet
\cite{le2015tiny}. Both the CIFAR-10 and CIFAR-100 datasets consist of 50,000
training images and 10,000 test images of size 32 $\times$ 32. The TinyImageNet
dataset is a subset of ImageNet \cite{deng2009imagenet} consisting of 200
classes each with 500 training and 100 test samples.

In line with standard practice, we used a mismatch ratio to designate the
proportion of known classes in the unlabeled dataset. Mismatch ratios of 20\%,
30\%, and 40\% represent varying degrees of openness across datasets. For
instance, a 20\% mismatch on CIFAR-10 means that 2 out of 10 classes are
considered known, while the remaining 8 classes are treated as unknown.
To train EKUS for 2D object classification, pseudo-unknown candidates are
selected from the unlabeled pool at each AL cycle by taking the top-$\rho=5\%$
samples with the highest energy values. During optimization, EKUS mini-batches
are constructed using equal numbers of labeled and pseudo-unknown samples.

For 3D object classification, we evaluated on ModelNet40 \cite{wu20153d}, which
contains 12,311 CAD models across 40 categories with a standard split of 9,843
training and 2,468 test shapes. We followed the conventional PointNet protocol
by sampling a fixed number of 1,024 points per shape, normalizing to the unit
sphere, and applying rotation and jitter augmentations. 
The 3D open-set splits follow the same mismatch-ratio definition as in the 2D
setting, and EKUS uses the same pseudo-unknown selection ratio with balanced
mini-batch construction during training.

\textbf{AL settings.} We initialized the labeled dataset by randomly sampling a
small fraction of samples from the known classes using 1\% of the known-class
samples for CIFAR-10, and 8\% for CIFAR-100 and TinyImageNet. Each experiment
was conducted over 10 AL cycles, where in each cycle 1,500 samples were
selected from the unlabeled pool for annotation. To ensure the reliability of
the results, all the experiments were repeated with three different random
initializations and the final results were averaged across these runs. For 3D,
we matched the relative annotation budget to the dataset scale. The initial
seed was set to 10\% of the training examples from the known classes, and each
cycle selected 300 samples from the remaining unlabeled pool. We ran the same
number of AL cycles as in 2D.

\textbf{Implementation details.} All 2D experiments utilized a ResNet-18
backbone for feature extraction. The model was trained using stochastic
gradient descent with an initial learning rate of 0.01, momentum of 0.9, and
weight decay of 0.0005. Each model was trained for 200 epochs per AL cycle with
a batch size of 128. In all experiments, the hyperparameters $\alpha$, $\beta$,
$\lambda$, and $\gamma$ were each set to a constant value of 0.1, 0.1, 0.2, and
0.2, respectively. For $L_{EKUS}$, the margin hyperparameters $\delta_k$ and
$\delta_u$ were set to $-23$ and $-5$, respectively, and $\delta_s$ in
$L_{ESS}$ was set to $-20$. In the 3D experiments, EKUS and ESS used PointNet
to process unordered point sets via shared MLP blocks and symmetric pooling. We
trained with Adam (learning rate 0.001 with cosine decay), batch size 32, and
200 epochs. The hyperparameters $\alpha$, $\beta$, $\lambda$, and $\gamma$ were
kept the same as in 2D. For the margin parameters, we used slightly shifted
values: $\delta_k=-10$, $\delta_u=-2$, and $\delta_s=-8$, as the energies
obtained with PointNet lie in a narrower range compared to ResNet. Please see
the appendix for a sensitivity analysis of the margin parameters.

\textbf{Baselines.} To validate the efficacy of our approach, we compared it
against several AL baselines including the following traditional and OSAL
methods. (1) Random: Random selection of samples from the unlabeled pool.  (2)
Entropy \cite{wang2014new}: Selection based on the predictive entropy of the
model. (3) MQ-Net \cite{park2022meta}. (4) LfOSA \cite{ning2022active}. (5)
BUAL \cite{zong2024bidirectional}. (6) EOAL \cite{safaei2024entropic}.  We
evaluated EB-OSAL against these competing methods in terms of object
classification accuracy. We adapted Random and Entropy on PointNet features for
3D object classification. To our knowledge, there are no established 3D OSAL
baselines, so we report these adapted baselines for reference.

\subsection{Results}
\vspace{-8mm}
\begin{figure*}
\centering
\setlength{\abovecaptionskip}{0.11cm}
\subfloat{
  \includegraphics[width=.32\textwidth]{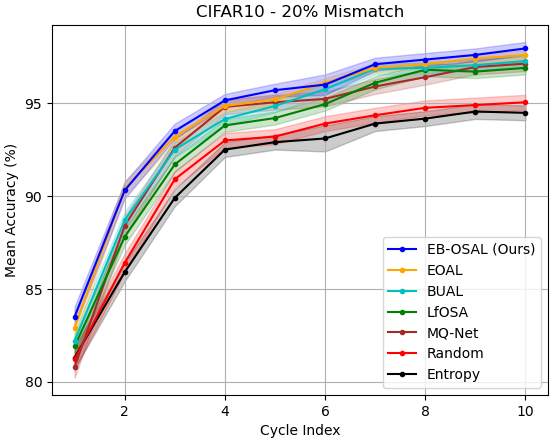}
  \label{subfig:cifar10_20}}
  \hfill
\subfloat{
  \includegraphics[width=.32\linewidth]{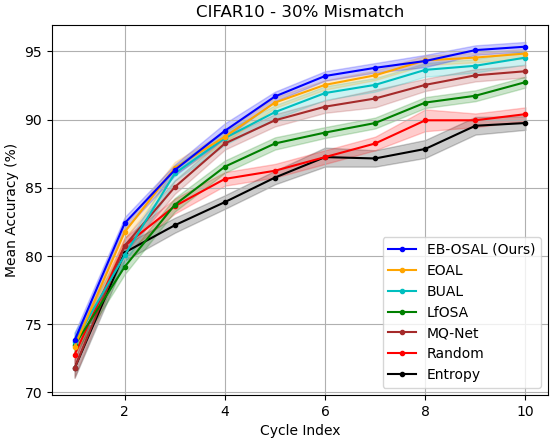}
  \label{subfig:cifar10_30}}
  \hfill
\subfloat{
  \includegraphics[width=.32\linewidth]{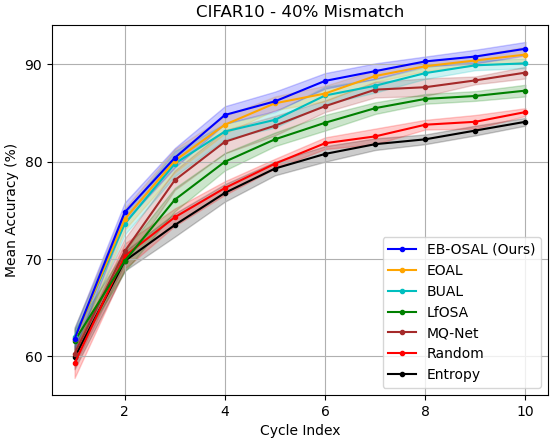}
  \label{subfig:cifar10_40}}\\
\caption{Object classification accuracy on CIFAR-10 with mismatch ratios of
20\%, 30\%, and 40\%.}
\label{fig:cifar_10}
\vspace{-12mm}
\end{figure*}

\begin{figure*}
\centering
\setlength{\abovecaptionskip}{0.11cm}
\subfloat{
  \includegraphics[width=.32\textwidth]{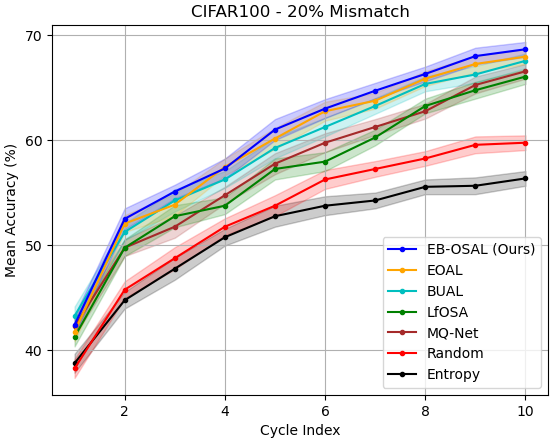}
  \label{subfig:cifar100_20}}
  \hfill
\subfloat{
  \includegraphics[width=.32\linewidth]{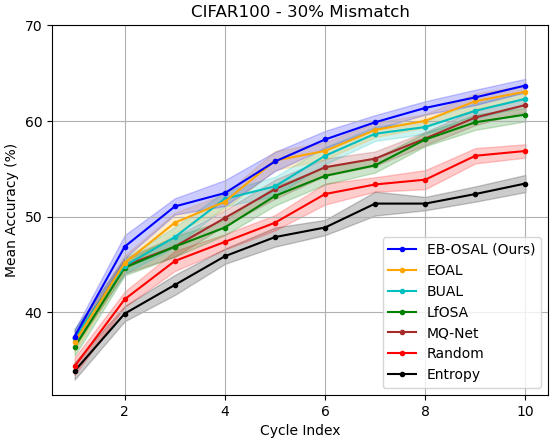}
  \label{subfig:cifar100_30}}
  \hfill
\subfloat{
  \includegraphics[width=.32\linewidth]{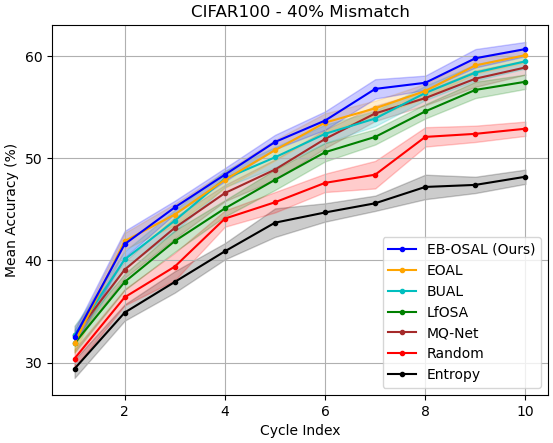}
  \label{subfig:cifar100_40}}\\
\caption{Object classification accuracy on CIFAR-100 with mismatch ratios of
20\%, 30\%, and 40\%.}
\label{fig:cifar_100}
\vspace{-12mm}
\end{figure*}

\begin{figure*}
\centering
\setlength{\abovecaptionskip}{0.11cm}
\subfloat{
  \includegraphics[width=.32\textwidth]{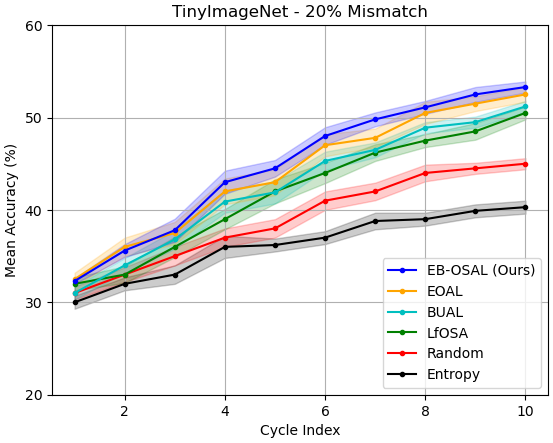}
  \label{subfig:tinyimagenet_20}}
  \hfill
\subfloat{
  \includegraphics[width=.32\linewidth]{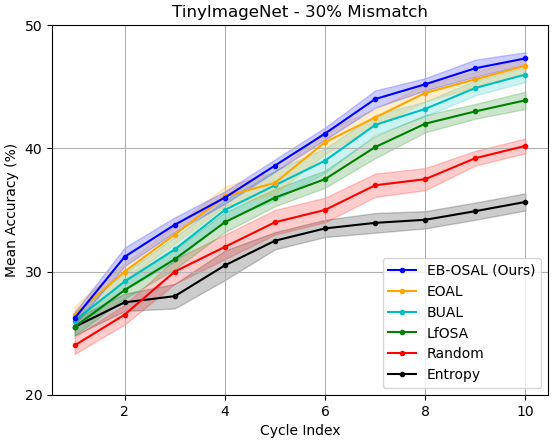}
  \label{subfig:tinyimagenet_30}}
  \hfill
\subfloat{
  \includegraphics[width=.32\linewidth]{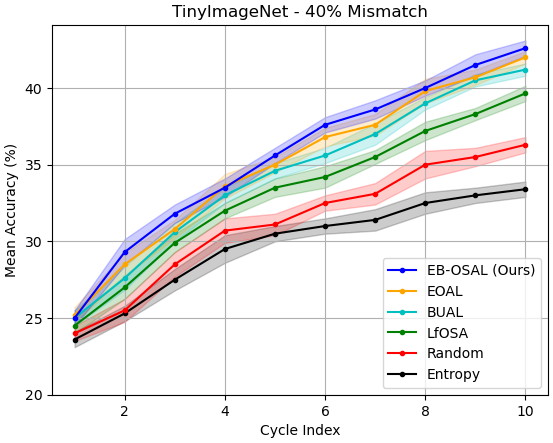}
  \label{subfig:tinyimagenet_40}}\\
\caption{Object classification accuracy on TinyImageNet with mismatch ratios of
20\%, 30\%, and 40\%.}
\label{fig:tinyimagenet}
\vspace{-5mm}
\end{figure*}

\begin{figure*}
\centering
\setlength{\abovecaptionskip}{0.11cm}
\subfloat{
  \includegraphics[width=.32\textwidth]{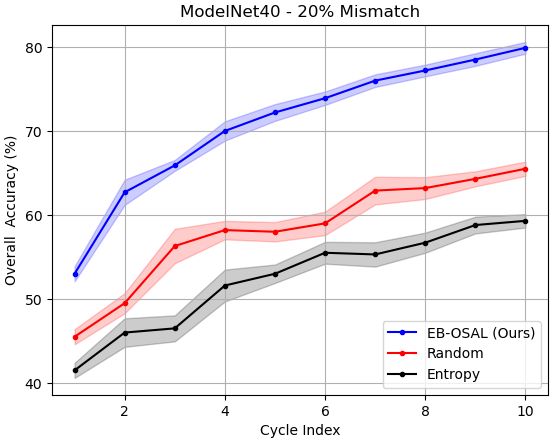}
  \label{subfig:modelnet40_20}}
  \hfill
\subfloat{
  \includegraphics[width=.32\linewidth]{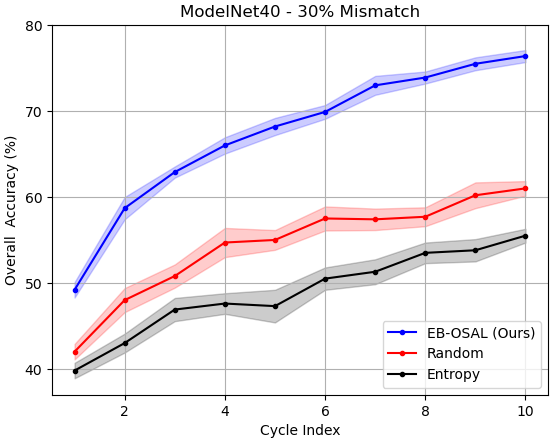}
  \label{subfig:modelnet40_30}}
  \hfill
\subfloat{
  \includegraphics[width=.32\linewidth]{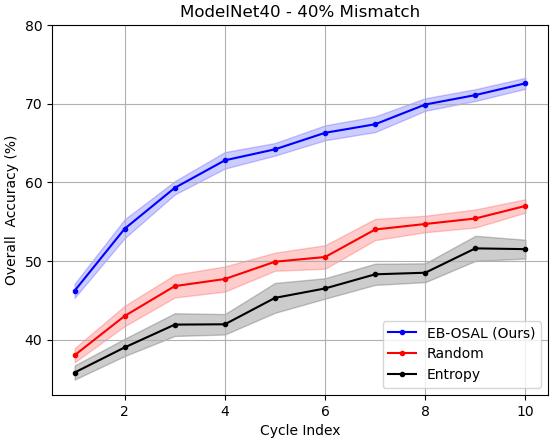}
  \label{subfig:modelnet40_40}}\\
\caption{Object classification overall accuracy on ModelNet40 with mismatch
ratios of 20\%, 30\%, and 40\%.}
\label{fig:modelnet40}
\vspace{-4mm}
\end{figure*}
Figs.~\ref{fig:cifar_10}-\ref{fig:tinyimagenet} present the classification
accuracy on CIFAR-10, CIFAR-100, and TinyImageNet across mismatch ratios of
20\%, 30\%, and 40\%. Our EB-OSAL framework consistently outperforms all
baselines across all datasets and mismatch ratios. The superior performance
demonstrates the efficacy of EKUS in filtering out unknown samples, combined
with ESS, which selects the most informative samples among the likely known
samples for labeling.

Furthermore, the results reveal that traditional AL methods, such as entropy,
can perform worse than random selection in open-set scenarios. This verifies
that traditional sampling strategies do not account for the presence of unknown
classes. In particular, they may mistakenly prioritize unknown samples, providing
little or no benefit for learning the known classes. The random selection
baseline, which selects samples without regard to informativeness, performs more
robustly in open-set settings than traditional AL methods, as it avoids the
over-selection of unknown samples. However, EB-OSAL significantly surpasses
random selection by focusing on samples that both belong to known classes and
are informative, indicating that our dual-stage approach effectively addresses
the limitations of conventional AL techniques in open-set conditions.

On ModelNet40, we report the overall classification accuracy across the same
mismatch ratios (Fig.~\ref{fig:modelnet40}). The trends mirror the 2D case:
EB-OSAL as a whole consistently outperforms the baseline methods. These overall
curves demonstrate the aggregate benefit of our dual-stage design. The specific
contributions of EKUS and ESS are further disentangled in an ablation study,
which shows how EKUS prevents unknown contamination and ESS provides additional
gains by prioritizing informative known-class shapes. An additional analysis on
their contributions and parameter sensitivity is provided in the appendix.

\section{Conclusion}
\label{sec:conclusion}
In this paper we introduced EB-OSAL, a novel energy-based framework for OSAL
with a dual-stage design integrating EKUS and ESS. EKUS filters out unknown
samples by assigning high-energy scores, effectively isolating relevant data.
ESS then prioritizes the retained known samples for annotation by combining
uncertainty and energy-based informativeness, ensuring that selected samples
contribute maximally to learning. Extensive experiments on CIFAR-10, CIFAR-100,
TinyImageNet, and ModelNet40 with varying mismatch ratios show that EB-OSAL
outperforms the latest OSAL benchmarks on both 2D and 3D object classification,
demonstrating robustness and adaptability in open-set conditions. Future work
will extend this methodology to other applications, including object detection
and segmentation, where handling uncertainty and unknowns is more challenging.

\bibliographystyle{splncs04}
\bibliography{references}

\section*{Appendix}
\appendix
This appendix provides additional experimental results and analyses to
complement the main paper. In particular, we report detailed ablation studies
that isolate the contributions of EKUS and ESS, and we present sensitivity
analyses of the margin hyperparameters $\delta_k$ and $\delta_u$ for both 2D
and 3D object classification tasks. These results further validate the
robustness and effectiveness of the proposed EB-OSAL framework.

\section{Ablation Study}
\begin{figure}
\centering
\setlength{\abovecaptionskip}{0.01cm}
\includegraphics[width=0.45\textwidth]{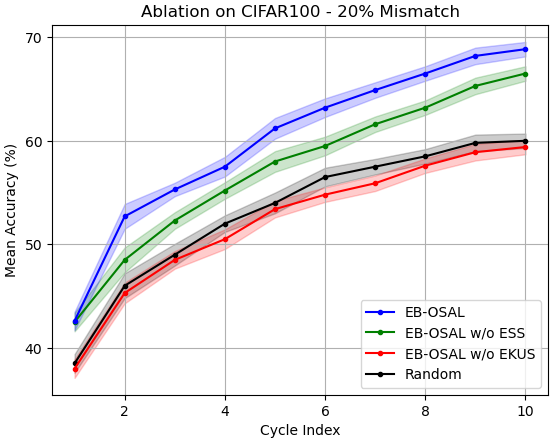}
\caption{An ablation study on CIFAR-100 with a mismatch ratio of 20\%.}
\label{fig:cifar100_ablation}
\end{figure}

To further investigate the contribution of each EB-OSAL component, we conducted
ablation experiments on CIFAR-100 with a 20\% mismatch ratio. This analysis
enables us to isolate the roles of EKUS and ESS and evaluate their individual
impact on the overall performance. The outcomes of the ablation study are
illustrated in Fig.~\ref{fig:cifar100_ablation}.

\textbf{EB-OSAL without EKUS.} This variant excludes the EKUS component,
meaning that no separation mechanism is used to filter unknown samples. Without
EKUS, the framework resembles a traditional AL setup, relying solely on ESS for
sample selection. However, without the known/unknown separation provided by
EKUS, the sample selection strategy becomes misleading in open-set conditions.
As a result, this variant performs worse than random selection, confirming that
a separation mechanism is crucial to avoid prioritizing unknown samples in
open-set scenarios.

\textbf{EB-OSAL without ESS.} In this variant, EKUS is retained to filter
unknown samples, but ESS is replaced with a default entropy-based selection
strategy. This setup performs significantly better than random selection and
the variant without EKUS, as the known/unknown separator ensures that only
likely known samples are considered for selection. The improvement highlights
EKUS's pivotal role in handling open-set conditions. However, the absence of
ESS limits the framework's ability to prioritize samples based on
informativeness, resulting in lower performance compared to the full OSAL
framework.

\begin{figure}
\centering
\setlength{\abovecaptionskip}{0.01cm}
\includegraphics[width=0.5\textwidth]{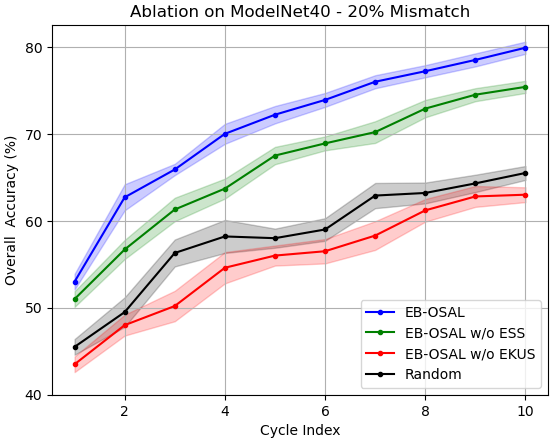}
\caption{An ablation study on ModelNet40 with a mismatch ration of 20\%.}
\label{fig:modelnet40_ablation}
\end{figure}

\textbf{ModelNet40 ablations.} We conducted the same ablation protocols on
ModelNet40. As shown in Fig.~\ref{fig:modelnet40_ablation}, the relative trends
match the 2D setting. Without EKUS, performance drops sharply due to severe
unknown contamination. Without ESS, EKUS + entropy selection remains robust,
yet underperforms the full model.

To summarize, the ablation results demonstrate that both EKUS and ESS play
essential roles in the EB-OSAL framework. EKUS provides critical filtering of
unknown samples, addressing the open-set nature of the data. ESS refines the
selection of known samples, ensuring that the most informative samples are
prioritized. The dual-stage design delivers the largest improvements, and the
combination of these two components enables EB-OSAL to achieve exceptional OSAL
performance.

\section{Hyperparameter Sensitivity Analysis}
\begin{figure}
\centering
\setlength{\abovecaptionskip}{0.05cm}
\includegraphics[width=0.6\textwidth]{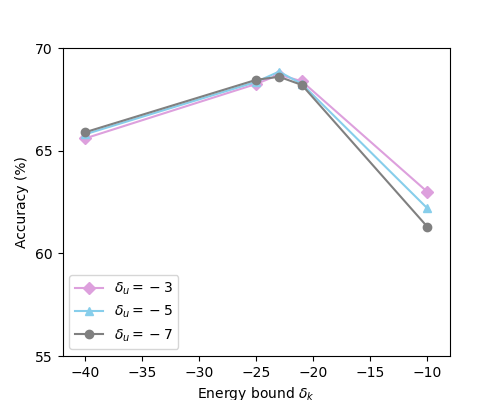}
\caption{The impact of margin parameters on CIFAR100.}
\label{fig:sensitivity_cifar100}
\end{figure}

\begin{figure}
\centering
\setlength{\abovecaptionskip}{0.05cm}
\includegraphics[width=0.6\textwidth]{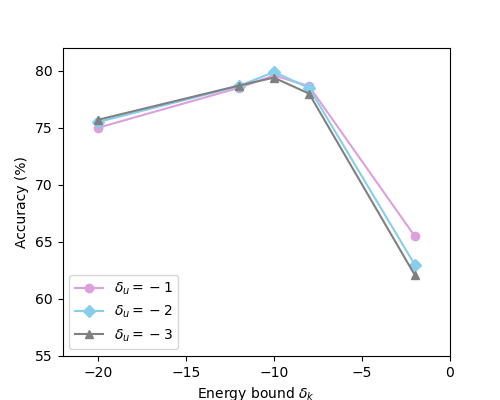}
\caption{The impact of margin parameters on ModelNet40.}
\label{fig:sensitivity_modelnet40}
\end{figure}

The key hyperparameters in the training objective are the margin parameters
$\delta_k$ and $\delta_u$, which enforce separation between known and unknown
samples when training EKUS (\eqref{eq:hinge} in the main paper). Specifically,
the model penalizes known samples with energy higher than $\delta_k$, and
unknown samples with energy lower than $\delta_u$. We assess the sensitivity of
these parameters on CIFAR-100 (2D) and ModelNet40 (3D), where each dataset is
evaluated under a 20\% mismatch ratio. For each dataset, we varied the margins
across a range of values and reported the final classification accuracy after
completing all AL cycles under these different parameter settings.

The sweep ranges were chosen to include both a central band of values near the
average energies of the known and unknown distributions, respectively, as well
as more extreme values to test robustness beyond this operating region.  For 2D
classification tasks (Fig.~\ref{fig:sensitivity_cifar100}), we vary $\delta_k
\in \{-10,-20,-23,-25,-40\}$ and $\delta_u \in \{-3,-5,-7\}$. For 3D
classification tasks (Fig.~\ref{fig:sensitivity_modelnet40}), we vary $\delta_k
\in \{-2,-8,-10,-12,-20\}$ and $\delta_u \in \{-1,-2,-3\}$.

The results demonstrate that EB-OSAL remains robust across a reasonable range
of margin values in both 2D and 3D settings. Performance degradation only
becomes noticeable when the gap between $\delta_k$ and $\delta_u$ becomes too
small, which limits EKUS's ability to effectively separate known and unknown
samples. Moreover, the degradation pattern is asymmetric. Setting $\delta_k$
too loosely (e.g., $-10$ in 2D, $-2$ in 3D) rapidly reduces performance due to
excessive contamination from unknowns, while setting it too strictly (e.g.,
$-40$ in 2D, $-20$ in 3D) causes only a gradual decline from over-penalizing
some knowns. Overall, the framework maintains strong performance within a
stable band of parameter values, confirming that EB-OSAL is not overly
sensitive to margin choices.

\end{document}